\documentclass{article}
\usepackage[square,numbers]{natbib}
\usepackage[utf8]{inputenc} 
\usepackage{amsmath}
\usepackage{algorithm}
\usepackage{algpseudocode}
\usepackage{amsfonts}       
\usepackage{graphicx}
\usepackage{textcomp}

\title{Deep Reinforcement Learning for Routing a Heterogeneous Fleet of Vehicles}

\author{Jos\'e Manuel Vera and Andres G. Abad \\ Industrial Artificial Intelligence (INARI) Research Lab \\
Escuela Superior Politecnica del Litoral}

\date{}
\begin{document}
\maketitle
\nocite{*}
\begin{abstract}
Motivated by the promising advances of deep-reinforcement learning (DRL) applied to cooperative multi-agent systems we propose a model and learning procedure to solve the \textit{Capacitated Multi-Vehicle Routing Problem} (CMVRP) with fixed fleet size. Our learning procedure follows a centralized-training and decentralized-execution paradigm. We empirically test our model and showed its capability for producing near-optimal solutions through cooperative actions. In large instances, our model generates better solutions than other commonly used heuristics. Additionally, our model can solve arbitrary instances of the CMVRP without requiring re-training.
\end{abstract}

\section{Introduction}

Given the impressive results of deep neural networks (DNN) in computer vision and natural language processing tasks, there has been recent interest in their incorporation to the reinforcement learning (RL) paradigm to tackle optimal control and sequential decision-making problems. The implementation of RL with DNN algorithms is referred to as deep-reinforcement learning (DRL) and has been recently used to solve combinatorial-optimization problems \cite{bengio}.

In this work, we aim to apply DRL to provide an end-to-end method to solve the \textit{Vehicle Routing Problem} (VRP) with multiple vehicles and heterogeneous capacities. We propose a model and a training procedure to route a fleet of vehicles with different capacities to act cooperatively and solve the routing problem.

\section{Related Work}

Most previous works on using neural networks to solve combinatorial-optimization problems as end-to-end methods formulate the problem, either, as a sequence of inputs or as a graph representation.

The pioneering work of \cite{vinyals} proposed an architecture called \textit{pointer networks} based on recurrent neural networks (RNN), that uses an attention mechanism, as in \cite{bahdanau}, to create pointers to a fixed set as outputs which allows to solve sequential combinatorial problems. Learning was achieved by maximizing the conditional probability of the training set in a supervised-learning manner. The training set consisted of a set of 2D points as inputs, and the solutions obtained from an approximate solver as labels; this method was applied for solving the \textit{Travelling Salesman Problem} (TSP). A drawback of this method was the high computational cost required for generating the training set.

In \cite{bello}, DRL was applied to solve the TSP using a \textit{pointer network} as the policy function and an auxiliary network called \textit{critic} to learn the expected tour length of an input sequence; training was achieved using the \textit{Advantage Actor Critic} (A2C) algorithm (see \cite{sutton}). More recently, \cite{nazari} applied a DRL model to produce near-optimal solutions for the VRP, generalizing the model in \cite{bello} by considering a dynamic system.

With respect to graph representations, \cite{dai} used a neural network architecture called \textit{structure2vec} to represent the problem instance as a latent space vector; together with RL, their method was applied to solve various combinatorial problems. This formulation, however, could not be applied to the VRP since it assumed that the graph is static through time. In \cite{kool}, an \textit{attention graph network} is used to represent the problem instances as vectors; the method produces competitive solutions for various combinatorial problems, including the VRP.

In this work, we extend the model proposed in \cite{nazari}. Because of its sequential nature and simplicity, it is useful for the formulation of the sequential decision making of multiple agents.

\section{Background}

\subsection{Capacitated Vehicle Routing Problem}
We consider an specific instance of the VRP in which $N$ vehicles, each with specific capacity, must deliver items to $M$ customers, each with finite specific demand. It is further assumed that all demands are smaller than the vehicle capacity. In order to satisfy the demand of each customer the vehicles must create routes starting and ending at a depot node. When the vehicle's load runs out, it returns to the depot to refill. The objective is to minimize the total route length of all vehicles while satisfying the demand of all customers. This problem can be termed the \textit{Fleet Size and Mix Vehicle Routing Problem} (FSMVRP) \cite{gheysens} with a fixed fleet size. We call it the \textit{Capacitated Multi-Vehicle Routing Problem} (CMVRP) but we will refer to it as the VRP throughout this work.

The mathematical programming formulation of this problem yields an exponential number of constraints with respect to the number of customers, making it computationally intractable for medium-to-large size problems.

\subsection{Sequence-to-Sequence Learning}

Each agent is sequentially given an input to make a decision at each timestep---the mechanism used to generate the decisions followed by each agent is its policy. Since decisions must be made sequentially, it seems natural to model this policy as a sequence-to-sequence model.

Given an input sequence $X_t=\{x_i\}_{i=0}^{t}$ the model finds the conditional probability of the output sequence $Y_t=\{y_i\}_{i=0}^{t}$ \cite{sutskever}. By assuming the Markov property, we can express this as
\begin{equation}
P(y_0,...,y_{T} | x_0,...,x_T ) = \prod_{t=0}^{T} P(y_{t+1} | Y_t, X_t  ).
\end{equation}
Recurrent neural networks are commonly used in sequence-to-sequence models to estimate this conditional probability.

\subsection{Attention Mechanism}
A sequence-to-sequence model assumes that the output sequence is formed by elements of a fixed set. Unlike the sequence-to-sequence model, the VRP solution (output) is a permutation of the problem nodes (input). To achieve this required behaviour we use a mechanism called \textit{attention} (see, for example, \cite{bahdanau}, \cite{vinyals}, and \cite{nazari}).

This technique is used to query information from all elements in the input-nodes set. To construct the output sequence, an affinity function is evaluated, with each node and the last output of the model, to generate a set of scalars. Then, by applying the \textit{softmax} function to these scalars, we obtain the attention given to each element of the input set at each timestep.

\section{Method}

\begin{figure*}
\centering
\includegraphics[scale=.5]{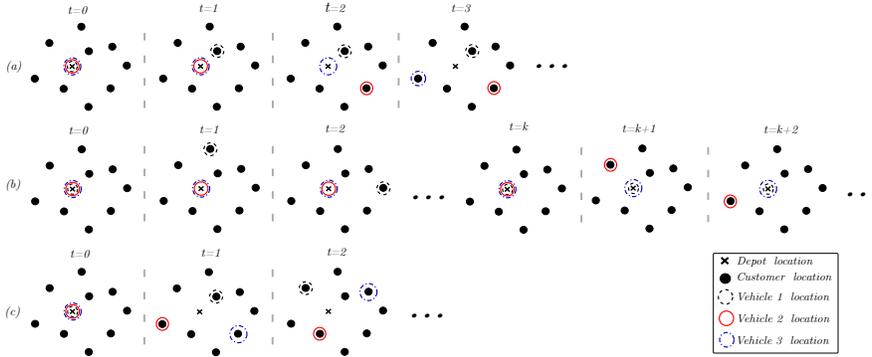}
\caption{Actions of a fleet of 3 vehicles following decision procedures (a) Case 1: Agents make single, sequential and alternating action at each timestep, (b) Case 2: Agents generate routes in a sequential and alternating order and (c) Case 3: Agents make simultaneous actions at each timestep}
\label{fig:casos}
\end{figure*}

Customer locations are considered on a 2D Euclidean space. Customers and depot locations are randomly generated in the unit square. It is assumed that the demand in each node, except the depot node, can take a discrete value uniformly distributed between $1$ and $9$. Throughout this section we will use the terms ``agent" and ``vehicle" interchangeably.

A problem instance $\mathcal{P}$ corresponds to a set of tuples
\begin{equation}
\mathcal{P} \doteq \{\textbf{s},\textbf{d},\textbf{l},\textbf{p}\},
\end{equation}
where
\begin{itemize}
    \item $\textbf{s} = \{s^i\}_{i=1}^M$ are the coordinates of customers;
    \item $\textbf{d} = \{d^i\}_{i=1}^M$ are the demand of customers;
    \item $\textbf{l} = \{l^j\}_{j=1}^N$ are the capacities of vehicles; and
    \item $\textbf{p} = \{p^j\}_{j=1}^N$ are the locations of vehicles.
\end{itemize}

The problem instance can be seen as the initial state of the problem. Agents will act on the problem  changing the original state.  The state of the problem at timestep $t$ is $X_t \doteq \{\, \textbf{s},\textbf{d}_t, \textbf{l}_t,\textbf{p}_t \, \}$.

In our formulation, agents act cooperatively to satisfy the demand of all customers. The policy of each agent is modeled with DNNs and trained using RL. The work of \cite{lowe} developed an algorithm---following the paradigm of centralize training and decentralize execution---to train agents in cooperative and competitive environments. Here each agent have their own policy which uses only local information at execution. Unlike \cite{lowe}, we propose a training procedure that allows our agents to access the information of all other agents and, thus, the state of the environment is the same for all agents.

To train the agents we have to specify the decision procedure that agents follow. This could be set in the following three cases.

\textbf{Case 1: Agents make single, sequential and alternating action at each timestep.}
In this procedure, the environment is stationary in the eyes of all agents and the VRP is formulated as a Markov Decision Process (MDP). This implies that each agent requires only information of the last state of the environment to make a decision and do not care about other agents actions (see Figure \ref{fig:casos} (a)). 

The policy of agent $j$ with parameters $\theta_j$ is
\begin{equation}\label{eq:policy}
\pi_{\theta_j}(a| \textbf{s}, \textbf{d}_t, \textbf{l}_t, \textbf{p}_t).
\end{equation}

The formulation of the problem as an MDP implies that each agent has a policy and the decisions are sequential and following the same order at each timestep.

This proposed decision procedure allows us to apply a policy gradient algorithm to train different agents with different characteristics.

\textbf{Case 2: Agents generate routes in a sequential and alternating order.}
This procedure also considers that the environment is stationary and that all agents have access to the same information. The vehicles are sorted by capacity in descending order. Each agent starts at the depot and make a series of actions until it returns to the depot; then the following agent have to make a series of actions and so on. When an agent is making a decision all other agents are in the depot so its policy does not need the information of positions and loads of the other agents. Thus, the state is defined as $X_t \doteq \{\, \textbf{s},\textbf{d}_t \, \}$.

The policy of agent $j$ with parameters $\theta_j$ is
\begin{equation}
\pi_{\theta_j}(a| \textbf{s}, \textbf{d}_t)\label{eq:policy_2}.
\end{equation}

Figure \ref{fig:casos} (b) depicts an example where the first vehicle started to generate a route until it returns to the depot at timestep $k$. Then the next vehicle must generate its route at timestep $k+1$.

\textbf{Case 3: Agents make simultaneous actions at each timestep.}
This procedure cannot model the problem as an MDP (see Figure \ref{fig:casos} (c)). If all agents make an action simultaneously the observation which is based the agent to make an action will change immediately after making an action, so the environment states are not useful to make decisions. Another issue is the difficulty of simultaneously imposing the restrictions of the problem to the actions of all agents.

For example, if a customer hasn't been visited by any agent then, in the next timestep is possible for all agents to visit this customer which is not a desirable behavior for the agents.


In what follows, we present a model and training algorithm for generating solutions for the VRP considering the decision procedure described in Case 1. In subsections A and B we will explain how policy in Equation (\ref{eq:policy}) is modeled and how the training is performed.

\subsection{Architecture}\label{sec:architecture}

Following A2C algorithm, we call the policy of each agent the \textit{actor network}. Additionally, this algorithm uses another neural network called the \textit{critic network}.

\begin{figure}[htbp]
\centerline{\includegraphics[scale=0.33]{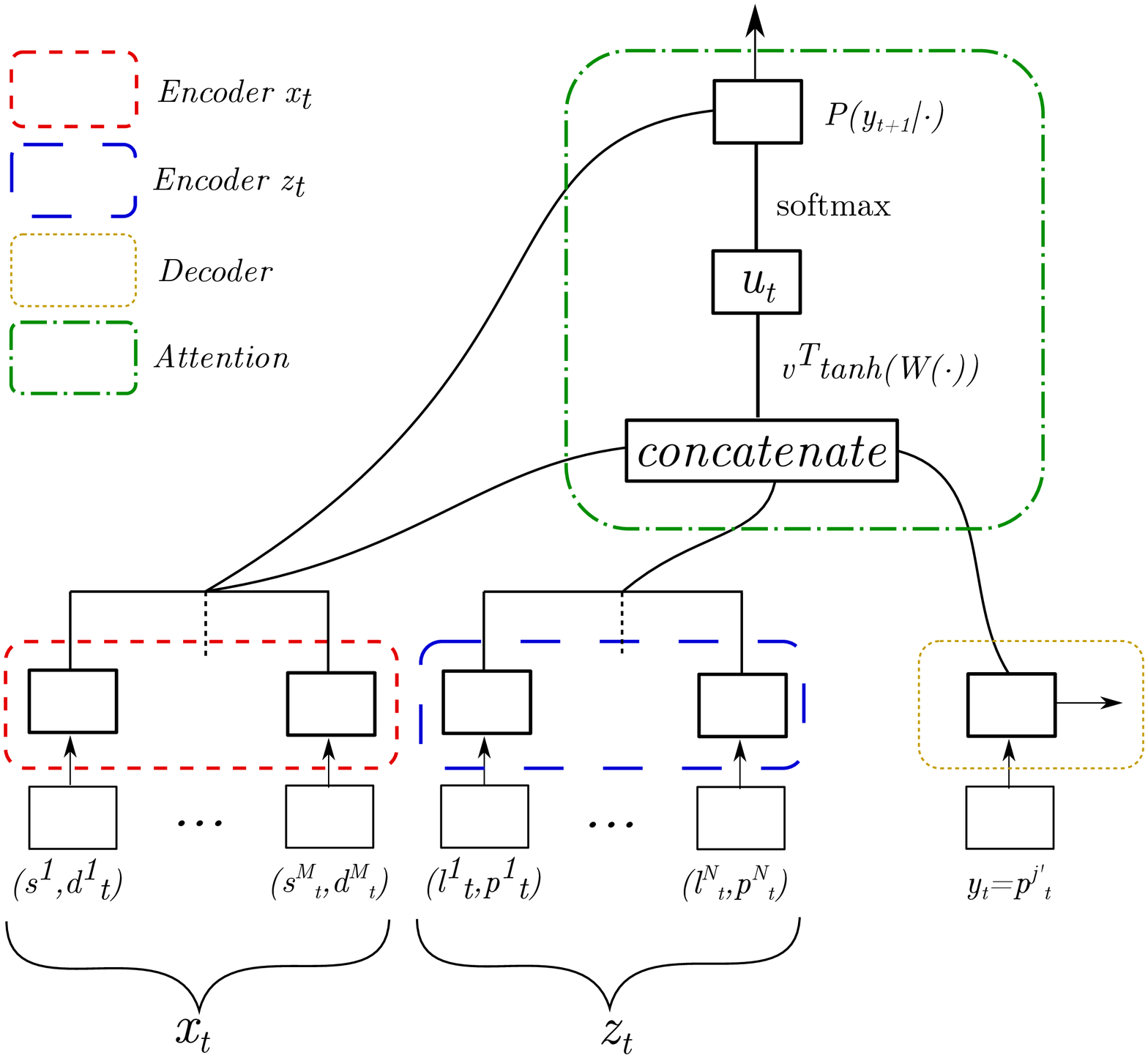}}
\caption{Proposed model for actor network (Case 1).}
\label{fig:actor_network}
\end{figure}

The actor network follows the architecture in \cite{nazari}. We expand on what data is feed into the encoder and also change how the attention mechanism is defined.  The actor network consists of a sequence-to-sequence model with an encoder, decoder and attention mechanism (Figure \ref{fig:actor_network}). At each timestep $t$ two inputs are given: $x_t \doteq \{(s^i,d^i_t)_{i}^{M}\}$, which contains the information about the customers; and $z_t \doteq \{(l^j_t,p^j_t)_{j}^{N}\}$, which contains the information about the agents.


These inputs are given to an encoder which embeds into latent space vectors. These embedded vectors are combined with the output $h_t$ of a decoder, to output $y_{t+1}$ that points to one of the elements of the input $\textbf{s} \in x_t$. Furthermore, $y_{t+1}$ is the input for the next timestep of the decoder. If the vehicle $j$ is taking an action at timestep $t$ then the input for the decoder $y_t$ is the action taken by the previous vehicle (more specifically, the position of the vehicle to make previous action). This process generates a sequence and ends when a terminating condition is satisfied, e.g., when a specific number of steps are completed.

In order to introduce the restrictions of the problem we use a masking procedure in the output of the actor networks which sets the log-probabilities of infeasible actions to  $-\infty$.

\subsubsection{Encoder}
It consists of a series of embeddings, each receiving two inputs: $x_t$ and $z_t$, at timestep $t$. Each input have its own encoder to produce a $D$-dimensional vector.

\subsubsection{Decoder}
It is an RNN that receives the vehicle position $y_t$ and maintains a hidden state $h_t \in \mathbb{R}^D$.

\subsubsection{Attention Mechanism} \label{sec:attention}
Let $\bar{x}_t= (\bar{s}^i_t, \bar{d}^i_t)$ and $\bar{z}_t=(\bar{l}^i_t, \bar{p}^i_t)$ be the embedded inputs of the problem instance and $h_t$ the hidden state of the decoder at timestep $t$. We concatenate the embedding vectors $\bar{x}_t$ and $\bar{z}_t$ with the hidden state of the decoder $h_t$ and do a linear transformation with the parameters $W$. We then apply a hyperbolic tangent function (tanh) and multiply with the vector $v^T$. Finally, we apply a softmax to the output. Thus, we compute an attention vector as follows:
\begin{equation}
u_t = v^T \text{tanh} (W[\bar{x}_t;\bar{z}_t;h_t]).
\end{equation}

Then, the conditional probability is defined as
\begin{equation}
  P(y_{t+1} | Y_t, X_t  )= \text{softmax}(u_t).
\end{equation}
The learnable parameters of the attention mechanism are $v$ and $W$. The agents make a greedy action according to this conditional probability.

\subsubsection{Critic}
The critic is a feed-forward neural network which receives as input the static elements $\textbf{s}$, and returns the estimated total reward of the problem instance.

\subsection{Training}\label{sec:training}

To train the networks we use a policy gradient method known as the A2C algorithm. This algorithm uses two DNNs as function approximators: one is called the \textit{actor network} that parameterizes the stochastic policy to predict a probability distribution over the next action at any given state; and the other is the \textit{critic network} that estimates the total reward for any problem instance. We parameterize a stochastic policy $\pi$ with parameters $\theta$ for the embedding, decoder, and attention mechanism. We iteratively improve the policy of each agent $j$ by estimating the gradient of the expected rewards $J(\pi_{\theta_j})$ with respect to the policy parameters, obtained by
\begin{equation}\label{eq:grad_policy}
  \begin{aligned}
  \nabla_{\theta_j}J(\pi_{\theta_j}) = \mathbb{E}_{\tau_j \sim \pi_{\theta_j}} \nabla_{\theta_j}\text{log } P(\tau_j | x;z; \theta_j)  \Big( R(\tau_j | \pi_{\theta_j}; x;z)- \\
  V_{\phi}(s_k) \Big) \\
 \approx \frac{1}{BT}\sum_{k=1}^{B} \sum_{t=0}^T \nabla_{\theta_j}\text{log } \pi_j(a^k_t | x^k_t;z^k_t; \theta_j)\Big(R(\tau^k_j|\pi_j ; x^k;z^k)-\\
 V_{\phi}(s_k) \Big). \\
  \end{aligned}
\end{equation}

Here $V_{\phi}(s_k)$ is the critic shared between agents and it estimates the total reward solely from the nodes locations of problem instance $k$, $s_k$; and $R(\tau^k|\pi;x^k;z^k)$ is the total reward of the tours $\tau^k = \bigcup_j^N \tau^k_j$ given policy $\pi = \bigcup_j^N \pi_{\theta_j}$ and problem instance $k$.

To reduce the variance in the gradients, we use the critic network. Thus, the gradient is scaled by the \textit{advantage}, which is the difference between the total and estimated reward of the problem instance. The critic network is improved via gradient descent according to
\begin{equation}\label{eq:grad_critic}
\nabla_{\phi}\frac{1}{B} \sum_{k=1}^{B} \Big(V_{\phi}(s_k)- R(\tau^k | \pi;x^k;z^k) \Big)^2.
\end{equation}
The update of the parameters is done following the Adam algorithm \cite{kingma}. The training procedure is described in Algorithm \ref{algo:1} where the generation of the tours of the agents is detailed as well as the learning of the parameters.

\begin{algorithm}
 \caption{A2C training for Routing a Heterogeneous Fleet of Vehicles (Case 1)}
	\begin{algorithmic}[1]
 	\State Initialize the actor network with parameters $\theta_j$ for agent $j$ and critic network with parameters $\phi$.
 	 \State Define number of timesteps $T$, batch size $B$ and vehicles capacities $l^j$ for $j \in \{1,...,N \}$.
        \For{$n$ iterations}
        \State Randomly sample a batch $\{(s^i,d^i_0)_{i=1}^M\}_{k=1}^B $, where $s^i \in [0,1] \times [0,1]$ and $d_0^i \in [1,9]$ for $i \in \{1,...,M \}$.
  		\For{$t$ from $1$ to $T \times N$}
  			\For{$j$ from $1$ to $N$}
				\State choose $a^j_t$ according to policy $\pi_{\theta_j}( \cdot | x_t;z_t;\theta_i)$
				\State observe new state $(x_{t+1},z_{t+1})$ according to transition function $f( \cdot | a^j_t,x^j_t)$
  			\EndFor
  		\EndFor
  	\For{$j$ from $1$ to $N$}
  		\State Compute $\nabla_{\theta_j}$ \Comment{As (\ref{eq:grad_policy})}

  		\State $\theta_j \gets$\textit{Adam} ($\theta_j$,$\nabla_{\theta_j}$)
  	\EndFor 
  	\State Compute $\nabla_{\phi} $ \Comment{As (\ref{eq:grad_critic})}
  	\State $\phi \gets$ \textit{Adam} ($\phi$,$\nabla_{\phi}$)
  \EndFor
	\end{algorithmic}
	\label{algo:1}
\end{algorithm}

\section{Experiments}
We performed the experiments described in Table \ref{tab1:experiments} over a test set of size 1000 and present the average tour lengths obtained. Note that testing Case 2 would have resulted in infeasible run times because it does not allows parallel generation of the tours used for training.

\begin{table}[htbp]
\caption{Experiments}
\begin{center}
\begin{tabular}{c c c c }
\hline
Name&No. Customers& No. Vehicles&Capacities\\\hline
VRP10 & 10 & 3 & 10, 15, 20\\
VRP20 & 20 & 3 & 20, 30, 35 \\
VRP50 & 50 & 3 & 60, 70, 80\\
VRP80 & 80 & 3 & 80, 100, 120\\\hline
\end{tabular}
\label{tab1:experiments}
\end{center}
\end{table}

The  purpose of this experiment is to test our representation of the problem and the proposed training procedure to analyze its results in comparison with other methods. For this reason, we avoid to tune the training parameters in order to sincere the results. The extensive research in deep learning has shown that, in general, leveraging computation translates in better results. In this way, we decide to perform a tractable number of iterations.

Our proposed method (which we will refer to as DRL) was compared with Google's OR-Tools \cite{ortools}, Clarke-Wright Savings Heuristic and Sweep Heuristic. We used a successive approximation approach \cite{juan} with the Clarke-Wright Savings Heuristic (CW) \cite{clarke}. With the Sweep Heuristic \cite{wren} we allowed a sequential generation of tours. Given the total capacity of vehicles with respect to the total possible demand of all customers we allow at most two tours per vehicle to encourage the use of all vehicles. This was done to avoid the exploit of the vehicle with the highest capacity which would have created a bottleneck.

We formulate the fleet size and mix vehicle routing problem (FSMVRP) \cite{gheysens} and found an optimal solution using the Gurobi solver \cite{gurobi}. OR-Tools and FSMVRP assume that each vehicle can make at most one tour, so the sum of capacities of the vehicles must be larger than the total demand. Given this restriction we must get a larger fleet so the algorithms could yield a feasible solution, i.e., having two vehicles of the same capacity is equivalent to having one vehicle making two tours. We were able to find the optimal solution only for the VRP10 experiment. For VRP20 and larger instances, the computation was too large to test on 1000 instances. For example, for an instance of VRP20 with three vehicles it took 2350 seconds to solve a single problem instance with a 10\% optimal gap.

Given the possibility of using heterogeneous fleet to the solve the VRP, we proposed different problem instances to test this capability.

\section{Results and conclusions}

Table \ref{tab2:results} shows the average tour lengths of the different methods along with our proposed method DRL. The DRL method generates, on average, shorter tour lengths than compared heuristics in large instances. Observe that as larger the instance, greater is the difference between the DRL method and the heuristics. Moreover, it shows that the generated tours of the DRL method had lower standard deviation compared to these heuristics. As for the tours generated by OR-Tools, these, on average, had shorter lengths compared to our proposed method. Note that the decision procedure of Case 1 introduces bias to our method by assuming that the optimal decisions must be in sequential and alternating order between vehicles. This limits the solution space to only the ones that follow this structure and could not be capable of generating better solutions than OR-Tools which has more liberty on how to generate solutions. Since the exact solution is intractable for large instances we do not present these values. 

\begin{table}[htbp]
\caption{Average tour length using different baselines over a test set of size 1000}
\centering
\scalebox{0.85}{
\begin{tabular}{c c c c c c c c c }\hline

Baseline & \multicolumn{2}{c}{VRP10}  & \multicolumn{2}{c}{VRP20} & \multicolumn{2}{c}{VRP50} &  \multicolumn{2}{c}{VRP80}  \\\cline{2-9}
& mean & std & mean & std & mean & std & mean &std\\\hline
DRL & 5.614  &  1.155 & \textbf{7.280} & 1.059 & \textbf{9.706} & 0.929 & \textbf{11.232} & 1.748\\
Sweep &  \textbf{5.510 } & 1.695 &  10.137 & 2.306 &  24.128 & 4.894& 37.164  & 4.626\\
CW & 6.884  & 1.628 & 12.181 & 2.438 & 27.351 & 3.691& 43.090 & 7.402 \\\hline
OR-Tools & 5.484  & 1.238 & 6.121 & 0.852 & 8.428 & 0.819 & 10.970 & 1.017 \\\hline \hline
Optimal & 5.087 & 0  & - & - &  -  & - & - & - \\\hline 
\end{tabular}
}
\label{tab2:results}
\end{table}

Our proposed method shows competitive run times compared to considered heuristics (Table \ref{tab3:runtimes}). OR-Tools presented  much longer run times in average compared with our proposed method. Gurobi run times are not presented for large instances due to their computational intractability.

\begin{table}[h]
\caption{Average run time (in seconds) using different baselines over a test set of size 1000}
\centering
\begin{tabular}{c c c c c c}\hline
 & DRL & Sweep & CW & OR-Tools & Gurobi\\\hline
VRP10 & 0.018 & 0.009 & 0.004& 0.017 & 6.697\\
VRP20 & 0.025 & 0.012 & 0.026 & 0.035 & -\\
VRP50& 0.102& 0.029 &0.315 & 0.118 & -\\
VRP80 & 0.168 & 0.078 & 0.468 & 0.293 & -\\\hline
\end{tabular}
\label{tab3:runtimes}
\end{table}

The results obtained showed the potential of DRL for generating better policies to solve the  problem of routing a fleet of vehicles with heterogeneous capacities and to automate this task by finding a global optimal policy that can be applied to arbitrary instances.

\section{Conclusions and Future Work}
We proposed a model and training procedure for finding near-optimal solutions to the problem of routing multiple vehicles with heterogeneous capacities. Our trained model generates better solutions than commonly used heuristics for large instances; falling short to, however, Google's OR-Tools. It is important to note that our proposed model finds policies that can be used to automate the task of routing a heterogeneous fleet for any configuration of nodes. This is a limitation of methods like OR-Tools that need to set up and solve each instance individually. Furthermore, our proposed method has competitive run times compared to other methods.


As future work we are interested in developing a training algorithm for Case 2 that allows parallelization. This will likely generate better solutions since this case is similar to the successive approximation approach. It would also be interesting to develop training algorithms following the decision procedure of Case 3 to allow several agents to make decisions simultaneously that are optimal globally.

\section*{Acknowledgment}

The authors would like to thank Tiendas Industriales Asociadas Sociedad Anonima (TIA S.A.), a leading grocery retailer in Ecuador, for providing necessary funding for this research effort.

\bibliography{ma_vrp_ref}

\providecommand{\bysame}{\leavevmode\hbox to3em{\hrulefill}\thinspace}
\providecommand{\MR}{\relax\ifhmode\unskip\space\fi MR }
\providecommand{\MRhref}[2]{%
  \href{http://www.ams.org/mathscinet-getitem?mr=#1}{#2}
}
\providecommand{\href}[2]{#2}
\begin{thebibliography}{10}

\bibitem{bahdanau}
Dzmitry Bahdanau, KyungHyun Cho, and Yoshua Bengio, \emph{Neural machine
  translation by jointly learning to align and translate}, arXiv preprint
  arXiv:1409.0473 (2014).

\bibitem{bello}
Irwan Bello, Hieu Pham, Quoc~V. Le, Mohammad Norouzi, and Samy Bengio,
  \emph{Neural combinatorial optimization with reinforcement learning}, arXiv
  preprint arXiv:1611.09940 (2016).

\bibitem{bengio}
Yoshua Bengio, Andrea Lodi, and Antoine Prouvost, \emph{Machine learning for
  combinatorial optimization: a methodological tour d’horizon}, arXiv
  preprint arXiv:1811.06128 (2018).

\bibitem{clarke}
G.~Clarke and J.~W. Wright, \emph{Scheduling of vehicles from a central depot
  to a number of delivery points}, Operations Research (1964).

\bibitem{dai}
Hanjun Dai, Elias~B. Khalil, Yuyu Zhang, Bistra Dilkina, and Le~Song,
  \emph{Learning combinatorial optimization algorithms over graphs}, 31st
  Conference on Neural Information Processing Systems (2017).

\bibitem{gheysens}
Filip Gheysens, Bruce Golden, and Arjang Assad, \emph{A comparison of
  techniques for solving the fleet size and mix vehicle routing problem}, OR
  Spectrum (1984).

\bibitem{gurobi}
LLC Gurobi~Optimization, \emph{Gurobi optimizer reference manual}, 2018.

\bibitem{ortools}
Google Inc., \emph{Google's optimization tools (or-tools)}, 2019.

\bibitem{juan}
Angel~A. Juan, Javier Faulin, Jose Caceres-Cruz, Barry~B. Barrios, and Enoc
  Martinez, \emph{A successive approximations method for the heteregeneous
  vehicle routing problem: analysing different fleet configurations}, European
  Journal of Industrial Engineering (2013).

\bibitem{kingma}
Diederik~P. Kingma and Jimmy~Lei Ba, \emph{Adam: A method for stochastic
  optimization}, International Conference on Machine Learning (2015).

\bibitem{kool}
Wouter Kool, Herke van Hoof, and Max Welling, \emph{Attention, learn to solve
  routing problems!}, 7th International Conference on Learning Representations
  (2019).

\bibitem{lowe}
Ryan Lowe, Yi~Wu, Aviv Tamar, Jean Harb, Pieter Abbeel, and Igor Mordatch,
  \emph{Multi-agent actor-critic for mixed cooperative-competitive
  environments}, arXiv preprint arXiv:1706.02275 (2017).

\bibitem{nazari}
Mohammadreza Nazari, Afshin Oroojlooy, Martin Tak\'a{$\check{c}$}, and
  Lawrence~V. Snyder, \emph{Reinforcement learning for solving the vehicle
  routing problem}, 32nd Conference on Neural Information Processing Systems
  (2018).

\bibitem{sutskever}
Ilya Sutskever, Oriol Vinyals, and Quoc~V. Le, \emph{Sequence to sequence
  learning with neural networks}, arXiv preprint arXiv:1409.3215v3 (2014).

\bibitem{sutton}
Richardh~S. Sutton and Andrew~G. Barto, \emph{Reinforcement learning: An
  introduction}, MIT Press, 2018.

\bibitem{vinyals}
Oriol Vinyals, Meire Fortunato, and Navdeep Jaitly, \emph{Pointer networks},
  29th Conference on Neural Information Processing Systems (2015).

\bibitem{wren}
Anthony Wren and Alan Holliday, \emph{Computer scheduling of vehicles from one
  or more depots to a number of delivery points}, Operational Research
  Quarterly (1970-1977) (1972).

\end{thebibliography}

\end{document}